\documentclass[12pt]{article}
\usepackage{amsmath}
\usepackage{graphicx}
\usepackage{natbib}
\usepackage{url} 

\newcommand{\blind}{0}

\usepackage{amsmath}
\usepackage{amssymb}
\usepackage{graphicx,psfrag,epsf}
\usepackage{enumerate}
\usepackage{natbib}
\usepackage{float}
 \usepackage{caption}
 \usepackage{adjustbox}
 \usepackage{tabularx}
 \usepackage{multirow}
\usepackage{url} 

\begin{document}

\def\spacingset#1{\renewcommand{\baselinestretch}%
{#1}\small\normalsize} \spacingset{1}

\if0\blind
{
  \title{\bf A Probit Tensor Factorization Model For Relational Learning}
  \author {Ye Liu\\
    Department of Statistics, North Carolina State University\\
    Rui Song \\
        Department of Statistics, North Carolina State University\\
      Wenbin Lu\\
      Department of Statistics, North Carolina State University\\
      Yanghua Xiao\\
    School of Computer Science, Fudan University}
  \maketitle
} \fi

\if1\blind
{
  \bigskip
  \bigskip
  \bigskip
  \begin{center}
    {\LARGE\bf A Probit Tensor Factorization Model For Relational Learning}
   \end{center}
  \medskip
} \fi

\bigskip
\begin{abstract}
With the proliferation of knowledge graphs, modeling data with complex multi-relational structure has gained increasing attention in the area of statistical relational learning. One of the most important goals of statistical relational learning is link prediction, i.e., predicting whether certain relations exist in the knowledge graph. A large number of models and algorithms have been proposed to perform link prediction, among which tensor factorization method has proven to achieve state-of-the-art performance in terms of computation efficiency and prediction accuracy. However, a common drawback of the existing tensor factorization models is that the missing relations and non-existing relations are treated in the same way, which results in a loss of information. To address this issue, we propose a binary tensor factorization model with probit link, which not only inherits the computation efficiency from the classic tensor factorization model but also accounts for the binary nature of relational data. Our proposed probit tensor factorization (PTF) model shows advantages in both the prediction accuracy and interpretability. 
\end{abstract}

\noindent%
\vfill

\newpage
\spacingset{1.5} 
\section{Introduction}
\label{sec:intro}

With the ever-growing computing technology, a huge volume of structured relational data have been generated in recent years. An example is the creation of knowledge graphs, which are graph-structured knowledge bases that store relations between entities. Notable projects are YAGO \citep{suchanek2007yago}, DBpedia \citep{auer2007dbpedia}, NELL \citep{carlson2010toward}, the Google Knowledge Graph \citep{singhal2012introducing} and so on. These graphs have become an important asset in many aspects.
For instance, the Google Knowledge Graph \citep{singhal2012introducing}, which currently contains 70 billion relational facts between different entities, provides structured information about how topics are linked to one another. Such information enables the search engine to provide more relevant results. Knowledge graphs have also been used in several specialized domains. 
In spite of their huge advantages in representing complex data, these graphs more often than not are very incomplete, noisy and sparse. Consequently, the construction and completion of multi-relational data is of great interest nowadays, where the latter task can be formalized as link prediction in relational learning.  

Unlike the majority of research in machine learning that assumes data are independently distributed, this independence assumption does not hold in relational data. In fact, we rely on the dependent relations between data to provide valuable information. The learning model is expected to  propagate information about different kinds of relations between a set of entities. 
Multi-relational data typically describes different types of relations between a set of entities. The entities can be words in natural language processing (NLP) applications, users in social networks, or disease in health record data. One common way to represent the relations between entities is to use triplets of the form (subject, predicate, object). The predicates of interest usually represent the semantic relationship between entities. A formal way to represent the data is to use a three-way binary tensor $\textbf{X}$ of size $N \times N \times K$ with $N$ as the number of entities and $K$ as the number of relationship types. The k-th slice of the tensor $\textbf{X}$, which is $X_k \in \{ -1, 1 \} ^{N\times N}$ contains information about the k-th relationship, with $x_{ijk}=1,-1$ denoting a valid/invalid relationship $k$ from entity $i$ to entity $j$. The tensor is usually sparse since most of the triplets are unobserved. 

Two different assumptions are widely used to interpret the unobserved triplets, including the closed world assumption (CWA) and the open world assumption (OWA). Under the CWA, unobserved triplets are assumed to be invalid, while under the OWA,  unobserved triplets can be either valid or invalid.  Majority of previous works used the CWA. They did not distinguish between invalid facts and missing facts, which are both encoded as 0. Such a representation will certainly reduce the prediction accuracy because the unobserved facts and invalid facts carry different information. 
Although most of the large-scale knowledge graphs only contain information for valid triplets, it is not hard to generate invalid samples by using side information. \cite{chang2014typed} pointed out that entity type can be used to dictate whether some entities can be legitimate arguments of a given predicate. 

Furthermore, most of the previous works on statistical relational learning assumes a Gaussian distribution for the relational data, which gives rise to many variants of the tensor factorization models \citep{nickel2011three,chang2014typed,drumond2016multi}, where the binary nature of the relational data is not taken into account.
Such an assumption is neither intuitive nor appropriate from a statistical perspective. 
Another limitation of these models is that the predictions are continuous scores, which do not have a probabilistic interpretation. 
A reasonable threshold needs to be manually chosen in order to achieve an accurate prediction.

To modify the Gaussian assumption on the relational data, \citeauthor{nickel2013logistic} (\citeyear{nickel2013logistic}) proposed a logistic tensor factorization model, RESCAL\_Logit, where each $x_{ijk}$'s is assumed to follow a Bernoulli distribution. This model makes the prediction procedure totally data-driven and improves the prediction accuracy. 
The authors used a quasi-Newton optimization, where some dense matrices need to be calculated in each iteration to get the gradients. This significantly slows down the model fitting and makes it less efficient compared to other tensor factorization models.  Besides, \citeauthor{krompass2013non} (\citeyear{krompass2013non}) extended the RESCAL model with an non-invalid constraint on the parameters.

In this paper, we propose the Probit Tensor Factorization (PTF) Model to address these problems. By imposing a probit link on the binary random variables $x_{ijk}$'s, we essentially assume that the binary value of each $x_{ijk}$ depends on the sign of a corresponding latent normally distributed random variable $Z_{ijk}$. The latent $Z_{ijk}$ follows different conditional distributions given $x_{ijk}$ is valid, invalid or missing. All the analyses utilize the
R package "ptf", which has been developed for fitting the Probit Tensor Factorization Model. The main contribution of this paper are summarized as follows:

First, a PTF Model is proposed to perform link prediction, which not only takes advantage of the computational efficiency from the Gaussian tensor factorization model but also enjoys the prediction accuracy and interpretability of the logistic tensor factorization model. Second, under the open world assumption (OWA), the unknown triplets are treated differently with known-to-be-invalid triplets, which is a more efficient and proper way of handling missing data. Third, experiments on the moderate-sized real-world datasets demonstrate that our model outperforms the state-of-the-art approaches.

The rest of the paper is organized as follows. First, we summarize several state-of-the-art relational learning models. Then we illustrate the framework of our model, with an analysis of computation complexity. After that, we compare prediction performance of our model with the current state-of-art models in real data applications. Finally, we provide a brief conclusion and discussion.

\section{Related Work}

In this section, the related work are organized into three parts. We start by reviewing related works in tensor factorization, then summarize related works in the area of relational learning. Finally, we connect our work to the existing network analysis literature.

\subsection{Related work in tensor factorization}

Tensors are generalizations of matrices to higher dimension and have long been studied. It is common to observe missing elements in a multi-dimensional tensor, which motivates the low-rank assumption. 
The earliest tensors and their decompositions can be traced back to \cite{hitchcock1927expression}.  During the last decades,  there has been a large number of researches on tensor factorization related topics targeting on a great variety of applications. 
For example, \cite{aidini20181} proposed a 1-bit tensor factorization and applied it to image analysis, 
\cite{ghadermarzy2018learning} generalize the 1-bit tensor factorization model and demonstrated its usage in  for context-aware recommender systems.

One of the most widely used tensor factorization is the CANDECOMP/PARAFAC (CP) proposed by \cite{harshman1994parafac}, which can be written as $X_k = AD_kB^T + E_k$, where $X_k$ is the k-th slice in the tensor whose dimension is $I$ by $J$; $A$ is an $I$ by $R$ factor loading matrix for Mode A and $B$ is a $J$ by $R$ loading matrix for Mode B; $D_k$ is an $R$ by $R$ diagonal matrix; $E_k$ is the k-th slice of the residual tensor.  
Based on the CP decomposition, \cite{rai2014scalable} proposed a Bayesian low-rank decomposition which is applicable to relational learning setting and achieved good performances.

Another well-known tensor factorization model is the Tucker decomposition \citep[][]{tucker1966some}. The main idea is to decompose a three-way tensor $\textbf{X} \in \mathcal{R}^{I \times J \times K}$ as 
$$ x_{ijk} = \sum_{p=1}^{P}\sum_{q=1}^{Q}\sum_{r=1}^{R} g_{pqr}a_{ip}b_{jq}c_{kr},$$
 for $ i = 1, ... , I, j = 1, ..., J, k = 1, ... K$.  
\cite{xu2013bayesian} proposed a InfTucker models, where binary data and missing data are taken into account.  

While these decomposition models have been applied to a great variety of areas \citep[][]{karami2012compression, rendle2010pairwise,  cong2015tensor}, the RESCAL model is a more suitable choice of tensor factorization for relational learning, because a more reasonable low-rank structure is used. More detail will be illustrated in the next subsection.

\subsection{Related work in relational learning}

Extensive methods have been explored to do link prediction with applications in all areas including knowledge graph completion \citep{wang2014knowledge,ji2016knowledge,nguyen2017novel}, biological network analysis \citep{almansoori2012link,sulaimany2018link, peng2020survey}, and social network analysis \citep{al2011survey, liben2007link, almansoori2012link}. One line of the research is the translation-based models. Among them, TransE \citep{bordes2013translating} is a very simple and effective one, where a relational fact is represented by a vector (h,r,t), which represents head, relationship, tail respectively. TransE assumes the score function $f_r(h,t)=\| h + r -t\|_2^2$, whose value is low if (h,r,t) holds and high otherwise. Later on, TransH \citep{wang2014knowledge} is proposed to overcome the flaws of TransE in dealing with 1-to-N, N-to-1 and N-to-N relation, where relations are modeled as a vector r on a hyperplane with a normal vector $w_r$. The score function is modified as $f_r(h,t)=\| h_{\bot} + r -t_{\bot}\|_2^2$, with $h_{\bot}=h-w_r^T h w_r$, 
$t_{\bot} = t-w_r^T t w_r$ and $\|w_r\|_2=1$. Another algorithm TransR \citep{lin2015learning} follows the same idea except that they embed entities and relations in distinct entity space and relation space. A more complicated model TATEC \citep{garcia2016combining} encompasses previous embedding works by combining two-way interactions with high-capacity three-way ones, which achieves even higher accuracy. Other variants of the translation-based embedding models including TransM \citep{fan2014transition}, TransA \citep{xiao2015transa},  STransE \citep{nguyen2016stranse} , and TransF \citep{feng2016knowledge}. While these translation-based models are effective and showed good performance, only the observed triplets with a set of sampled unobserved triplets are taken into account during optimization. 

Another important class of models represent relational facts as tensors, which draws a growing interest due to their natural representation of multi-relational data and its ability to utilize all the triplets' information in the data. 
\cite{franz2009triplerank} proposed the TripleRank method, where they used PARAFAC tensor factorization to perform link prediction. The data is represented by a tensor $\textbf{X} \in \mathbb{R}^{k \times \ell \times m}$
, which is decomposed using $\textbf{X} = \sum_{k=1}^n \lambda_k \cdot U_1^k \circ U_2^k \circ U_3^k$, such that $U_1 \in \mathbb{R}^{k \times n}$, 
$U_2 \in \mathbb{R}^{\ell \times n}$,
$U_3 \in \mathbb{R}^{m \times n}$.
and $``\circ''$ denotes outer product. \cite{bader2007temporal} used a family of models called DEDICOM from the psychometrics literature to do tensor factorization in social networks, which shows the advantages of tensor factorization in dealing with such relational datasets. 

\cite{nickel2011three} proposed the RESCAL model that modified the DEDICOM by imposing a different constraint, which is more appropriate to use for multi-relational data. Specifically the model is $\textbf{X}_k \approx AW_kA^T$ where  $A$ contains the latent-component representation of the entities and $W_k$ models the interaction of the latent component in the $k$-th relationship. RESCAL in  \cite{nickel2011three} uses the ASALSAN algorithm to find optimal parameters, where matrix $A$ and matrices $W_k, k=1,2,...,K$ are updated using efficient alternating least squares (ALS). The RESCAL model demonstrated its scalability by conducting experiments on YAGO 2 ontology data, which consists of 2.6 million entities, 33 million facts and 87 predicates.  While the RESCAL achieves good performance in efficiency, scalability and accuracy, binary relational data are considered to be continuous in the model. As a result, an appropriate threshold is needed to be given before link prediction. To address this issue, Nickel et al. (2013) proposed a logistic tensor factorization methods. In this model, they assume each element in $X_k$ follows a Bernoulli distribution, which outperforms the original RESCAL model in terms of prediction accuracy measured by AUC. But a serious drawback of the logistic tensor factorization methods is its poor scalability where dense matrices $AW_kA^T$ needs to be computed in each iteration, whose complexity is quadratic in the number of entities. So it would be infeasible to use logistic tensor factorization in knowledge base like YAGO2. Some other extensions of RESCAL model that take the binary nature of the data into account have been explored by \cite{london2013multi} and \cite{hu2016topic}, \cite{ghadermarzy2018learning}, where the first one discussed different loss functions and used L-BFGS to estimate the optimal parameters. 
and the latter two are fully bayesian based factorization models.

Our proposed PTF model is an improvement on both the RESCAL model and logistic tensor factorization models in that it inherits the computation efficiency from the former and enjoys the prediction accuracy from the latter.

\subsection{Related work in network analysis}

Our work is also closely related with researches of network analysis in statistics literature, especially the literatures where the data is represented as a multi-way tensor data.  Two common multi-way tensor related topics have been studied extensively. 

One important topic in the area is to describe the high-dimensional multi-way tensor data with some low-rank representations.  
A commonly adopted frameworks is a model of the form $\textbf{X} =\textbf{M} + \textbf{E}$,  where $\textbf{X}$ denotes the observed tensor data, $\textbf{E}$ is noise and $\textbf{M}$ is a mean array that is often assumed to be of low rank and can characterize the signal of interest with fewer parameters. 
To infer the low-dimensional representation of the connection patterns $\textbf{M}$, different types of models have been explored  \citep{westveld2011mixed, hoff2011hierarchical, schein2015bayesian, minhas2016new, hoff2016equivariant,vervliet2019exploiting}. 

The other closely related line of research focuses on dynamic networks, which depict how the network data evolves over time.  The goal is to predict future networks and capture the linkage evolution.
It has been posited as a major topic of interest in many areas. 
A great number of researches within statistical frameworks have been proposed \citep{sarkar2005dynamic, sarkar2007latent, foulds2011dynamic, xu2014dynamic, durante2014nonparametric, sewell2015latent, hoff2011hierarchical,hoff2015multilinear},
where different hidden structures are assumed. 
Despite advances in statistical methods of dynamic network analysis, deep models with the advantage in modelling non-linear transformation of network structure have also been explored \citep{perozzi2014deepwalk,wang2016structural,grover2016node2vec,trivedi2017know}.
In this paper, while we are also modeling with three-way tensors, we specifically target at the task of missing link prediction and restrict to the third slice in the tensors to represent different types of relations between entities. 
We expect that for different types of relations, 
the low-rank representations of the entities will be interacted in different ways, which will be characterized by relation-specific latent matrices in our model.

\section{Model Description} 

Tensors are denoted by bold upper case letters, such as  $\textbf{X}$, $\textbf{Z}$ and $\textbf{W}$. Denote the k-th slice of a tensor $\textbf{Z}$ by $Z_k$,  the element in the $i$-th row, $j$-th column of $Z_k$ by $z_{ijk}$. Matrices are denoted by capital letters such as $A$, whereas $a_i$ denotes $i$-th row of matrix $A$.

\subsection{Model Framework}
Assume that there are K types of relationships between N entities. The relational facts can be represented as a three-way tensor $\textbf{X} \in {\{-1, 1\}}^ {N\times N \times K}$ with some missing values. 
$X_{k}$ is an asymmetric $N \times N$ matrix representing the facts in the k-th type of relation. Specifically, $x_{ijk}=1$ and $x_{ijk}=-1$ stand for existing (valid) and non-existing (invalid) triplet ($i$-th entity, k-th relationship, $j$-th entity) respectively. The goal of the proposed model is to estimate $Pr(x_{ijk}=1)$ for all missing triplets given the observed relations.  

A latent three-way tensor variable $\textbf{Z}$ is introduced with the same dimension as $\textbf{X}$, where each element $z_{ijk}$ follows a Gaussian distribution. More specifically, for the $k$-th relation slice, $$Z_k=A W_k A^T + \epsilon_k$$
where $\textbf{vec}(\epsilon_k)\sim N(0,I)$, where $A$ is a $N \times r$ matrix and $W_k$ is a $r \times r$ matrix.  Here $r$ is the factorization rank, $\textbf{vec}$ stands for the vectorization operator and $I$ is the identity matrix. The $i$-th row of $A$ can be interpreted as the latent component representation for the $i$-th entity, and $W_k$ can be viewed as the interaction effects of latent components for the k-th relation.  We expect that for different relations, the latent representations for entities will interact with each other in different ways, which can be captured by the different in $W_k$.
Note that the bilinear product $AW_kA^T$ can be written as $(AH)H^{-1}W_kH^{-T}(AH)^T$ for any invertible $R\times R$ matrix $H$. While $A$ and $W_k$ are not identifiable, the underlying score for a triplet is unchanged when $(A, W_k)$ is replaced by $(AH, H^{-1}W_kH^{-T})$.

A constraint that $x_{ijk}=sign(z_{ijk})$ for all $i,j,k$ is imposed, which implies the probit link so that $Pr(x_{ijk}=1) = Pr(z_{ijk}>0)$. 
There are two reasons for using the probit link based on the latent tensor variable $\textbf{Z}$. 
First, the model outputs probabilities for each of the unknown triplets to be valid, which allows interpretable, probabilistic inference.
Second, the use of the probit link enables an efficient iterative estimation method.

\subsection{EM Algorithm}

The basic idea of our model fitting is to maximize the log-likelihood, which is equivalent to minimizing cross-entropy loss.  Since a latent variable is included in our model, directly estimating of model parameter would be numerically intensive. We use the classic EM algorithm to handle the latent variable. 

EM algorithm is an iterative method to perform parameter estimation involving latent variables. It was first formally presented by \cite{dempster1977maximum} and has been a popular tool in numerical optimization. It consists of a E-step and a M-step.

\subsubsection{E-step} The complete data log-likelihood, where we assume the latent variable $\textbf{Z}$ is observed, can be written as $$log L(A, \textbf{W} | \textbf{X},\textbf{Z})=C - 0.5\sum_{ijk} (z_{ijk}-a_i^T W_k a_j)^2,$$
 where C is a constant. In addition, we impose the constraint such that $sign(z_{ijk})=x_{ijk}$ for those observed $x_{ijk}$, i.e. $z_{ijk}$ should be valid if $x_{ijk}=1$ and invalid if $x_{ijk}=-1$.

Then, the Q function can be written as
\begin{align*}
Q=&E_{A^{(t)},\textbf{W}^{(t)}}\{log L(A,\textbf{W}|\textbf{X},\textbf{Z})|\textbf{X}\}\\
=&E_{A^{(t)},\textbf{W}^{(t)}}\{C - 0.5\sum_{ijk} (z_{ijk}-a_i^T W_k a_j)^2 | \textbf{X} \} \\ 
=&- 0.5  \sum_{ijk} \left \{ [E_{A^{(t)},\textbf{W}^{(t)}}(z_{ijk}|x_{ijk})-a_i^T W_k a_j]^2  + E_{A^{(t)},\textbf{W}^{(t)}}(z_{ijk}^2|x_{ijk})-E_{A^{(t)},\textbf{W}^{(t)}}^2(z_{ijk}|x_{ijk})\right \} + C\\ 
=&-0.5\sum_{ijk}[E_{A^{(t)},\textbf{W}^{(t)}}(z_{ijk}|x_{ijk})-a_i^T W_k a_j]^2 +C_2+C,
    \end{align*}
where $C_2$ is another constant such that 
$$C_2 = -0.5[E_{A^{(t)},\textbf{W}^{(t)}}(z_{ijk}^2|x_{ijk})-E_{A^{(t)},\textbf{W}^{(t)}}^2(z_{ijk}|x_{ijk})],$$ 
and $E_{A^{(t)},\textbf{W}^{(t)}}$ denotes the expectation at the parameter values from last iteration t.

So in the M step, it is equivalent to minimize  
$\sum_{k=1}^K \|E_{A^{(t)},\textbf{W}^{(t)}}(Z_k|X_k)-AW_kA^T\|_F^2$.
Since the latent variable is assumed to follow a Gaussian distribution with constraint $x_{ijk}=sign(z_{ijk})$, we have at iteration t such that 
\begin{equation*} 
	z_{ijk}|x_{ijk}  
	\begin{cases}
                    \sim TN(\mu_{ijk},1; z_{ijk} \in (0, \infty)), if \; x_{ijk} = 1,  \\
                    \sim TN(\mu_{ijk},1; z_{ijk} \in (-\infty, 0)), if\; x_{ijk} = -1, \\
            	\sim N(\mu_{ijk},1) , if\;x_{ijk}\; is\;missing,
                 \end{cases}
\end{equation*} 
where $\mu_{ijk}=a_i^T W_k a_j$, and $a_i$, $a_j$ are $i$-th and $j$-th row from matrix A from last iteration. TN denotes truncated Gaussian distribution, with parameter space truncated to the domain defined after semicolon.
Then, the expectation of the truncated Gaussian distribution can be computed as
 \begin{equation*} 
	E_{A^{(t)},\textbf{W}^{(t)}}(z_{ijk}|x_{ijk})  =
	\begin{cases}
	          \mu_{ijk}+\frac{\phi(-\mu_{ijk})}{ 1-\Phi(-\mu_{ijk})},if\; x_{ijk} = 1, \\
                    \mu_{ijk}-\frac{\phi(-\mu_{ijk})}{\Phi(-\mu_{ijk})},if\; x_{ijk} = -1,  \\
                    \mu_{ijk}, \;if\; x_{ijk}\;missing,
                 \end{cases}
\end{equation*} 

For simplicity of the notations, we denote the expectation matrix $E_{A^{(t)},\textbf{W}^{(t)}}(Z_k|X_k)$ by $E_k$. For k=1,...K, we define $M_k$ as sparse matrices such that 
\begin{equation*} 
	M_{ijk}=
	\begin{cases}
	           \frac{\phi(-\mu_{ijk})}{ 1-\Phi(-\mu_{ijk})}, if\; x_{ijk} = 1,\\
                    -\frac{\phi(-\mu_{ijk})}{\Phi(-\mu_{ijk})}, if \; x_{ijk} = -1,  \\
            	  0, if\;x_{ijk}\; is \;missing,
                 \end{cases}
\end{equation*} 
where $\phi(.)$ and $\Phi(.)$ are the probability density function and cumulative distribution function for the standard Gaussian distribution.

Then we have $E_k=AW_kA^T + M_k$. At each iteration, we only need to compute the sparse matrices $M_k$'s rather than the dense matrices $E_k$'s. This technique will be illustrated in the M-step, which significantly saves the computation cost of the algorithm.  
 
\subsubsection{M-step} 
In the M-step, our goal is to maximize Q function, which is equivalent to solve the following optimization problem
$$\underset{A,\textbf{W}}{\min} \sum_{k=1}^K \|E_k-AW_kA^T\|_F^2. $$	

If $E_k$'s were calculated in the E-step, then the update for $A$ and $\textbf{W}$ would be the same as RESCAL model. However, we only need the matrices $M_k$'s to carry out the M-step. Most of the elements in $M_k$'s are usually zero, since  most relational datasets are sparse, i.e. only a small fraction of relational facts are known.
Not having to evaluate $E_k$'s saves us  $O(N^2 r)$ computation cost in each iteration. The parameter updating rules are based on the alternating least-squares (ALS) algorithm as follows. 

\textbf{Update W in M Step}:

When $A$ is fixed, our objective function about $W_k$ is
\begin{align*} 
f(W_k)&=\|E_k-AW_kA^T\|_F^2 \\
&=\|\textbf{vec}(E_k) - (A\otimes A)\textbf{vec}(W_k)\|^2,
\end{align*}
which becomes a least square problem.

Let $\beta = A\otimes A$, where $\otimes$ is the Kronecker product.

Then the update rule is
 \begin{align*}
\textbf{vec}(W_{k}) &\leftarrow (\beta^T\beta)^{-1}\beta^T\textbf{vec}(E_k) \nonumber\\
&=(\beta^T\beta)^{-1}\beta^T\textbf{vec}(A W_k A^T+M_k) \nonumber\\
&=(\beta^T\beta)^{-1}\beta^T(A\otimes A) \textbf{vec}(W_k) + (\beta^T\beta)^{-1}\beta^T\textbf{vec}(M_k) \nonumber\\
&=(\beta^T\beta)^{-1}\beta^T\beta \textbf{vec}(W_k) + (\beta^T\beta)^{-1}\beta^T\textbf{vec}(M_k)\\
&=\textbf{vec} (W_k)+(\beta^T\beta)^{-1}\beta^T\textbf{vec}(M_k).
   \end{align*}

In the update for $W_k$, the most expensive computation lies in computing $(\beta^T\beta)^{-1}\beta^T$. To reduce the complexity, To reduce the complexity, we apply the singular value decomposition(SVD) to $A$  as suggested by Chang et.al (2014), i.e. $A=U\Sigma V^T$ with $U$ and $V$ being 
orthogonal matrices and $\Sigma$ being a diagonal matrix. Then,
\begin{equation*}
(\beta^T\beta)^{-1} = (V \otimes V) (\Sigma^2 \otimes \Sigma^2)^{-1}(V\otimes V)^T.
\end{equation*}

The update rule can be rewritten as
\begin{align*}
\textbf{vec}(W_{k}) &\leftarrow \textbf{vec} (W_k)+(\beta^T\beta)^{-1}\beta^T\textbf{vec}(M_k) \\
&=\textbf{vec} (W_k)+(V \otimes V) (\Sigma^2 \otimes \Sigma^2)^{-1} (V\otimes V)^T (A\otimes A)^T\textbf{vec}(M_k) \\
&=\textbf{vec} (W_k)+ (V \otimes V) (\Sigma^2 \otimes \Sigma^2)^{-1} \textbf{vec}(V^T A^T M_k A V) \\
&= \textbf{vec} (W_k)+\textbf{vec}(V (\eta_k ./ \alpha) V^T),
\end{align*}
where $\eta_k=V^T A^T M_k A V = V^T (\beta^T \textbf{vec}(M_k)) V$, $\alpha=\text{diag}(\Sigma^2)\text{diag}(\Sigma^2)^T$ and $./$ denote element-wise division.

\textbf{Update $A$ in M Step}:

In order to simultaneously solve for the left matrix and right matrix $A$, we stack the data side by side as suggested by ASALSAN in \cite{bader2007temporal}.  Let $\bar{E} = (E_1,E_1^T, \;\;... \;\;,E_K ,E_K^T)$
, $\bar{W}=(W_1, W_1^T ,\;\;...\;\; W_K, W_K^T)$, then it is equivalent to write $\bar{E} = A \bar{W}(\textbf{I}_{2k} \otimes A^T)$. By setting its gradient to 0, we can get the update rule of $A$ as
\begin{align*} 
A \leftarrow &    \left\{   \overset{K}{\underset{k=1}{\sum}}(E_kAW_k^T +E_k^TAW_k)\right\} \left\{ \overset{K}{\underset{k=1} {\sum}}(B_k+C_k) \right\} ^{-1},
\end{align*}

where  $$B_k=W_kA^TAW_k^T,$$ $$C_k=W_k^TA^TAW_k.$$ 

We can compute $E_kAW_k^T$ and $E_k^T AW_k$ in the following way.
\begin{align*}
E_kAW_k^T& = (AW_kA^T+M_k)AW_k^T \nonumber\\
=&AW_kA^TAW_k^T+M_kAW_k^T \nonumber \\
=&AB_k+M_kAW_k^T,
\end{align*}
  Similarly, 
\begin{align*}
E_k^TAW_k = A(AW_k)^TAW_k+M_k^TAW_k = AC_k+M_k^TAW_k
\end{align*}

\subsubsection{Parameter Initialization}

Note that the objective function is non-convex. The optimization process searches for locally optimal solutions and is not guaranteed to converge to a global optimal. Initialization is crucial for the performance. Similar to RESCAL \citep{nickel2011three}, the initialization process is performed with a singular value decomposition over $\bar{X} = \sum_k (X_k + X_k^T)$ such that $\bar{X}  = U\Sigma V^T$. The initial value of $A$ is set to $U$. $W$ is randomly initialized with each element independently drawn from a standard normal distribution. We find such a initialization work well during our simulation study and real data applications.




\subsection{Complexity Analysis}

To ensure the algorithm is tractable, it is important to study the computation complexity for the learning algorithm. Recall that $N$ denotes the number of entities, $K$ denotes the number of relationship types, $r$ denotes the rank for the factorization. Let $T$ denote the number of known triplets. The total number of triplets is $N^2 K$. When only a small fraction of the triplets are known to be valid / invalid, $T$ is much smaller than $N^2 K$ and the matrices $M_k$'s in our algorithm are sparse matrices with only $T$ non-zero elements. 

The computation complexity is $O(Tr^2)$  for each E-step and $O(NKr^2 + TNr+Kr^3)$ for each M-step, which includes $O(NKr^2 + Tr+Kr^3)$ for updating $A$ and $O(Nr^2 + TNr+r^3)$ for updating $\textbf{W}$. In total, for each EM iteration, we need $O(NKr^2 + TNr + Kr^3)$ computation, which is linear in $N$ and $K$, $T$ and cubic in $r$.

In terms of memory complexity, the tensors and matrices we need to keep in memory include a sparse three-way data $\textbf{X}$ of size $N\times N \times K$ with $T$ non-zero elements, whose memory complexity is $O(T)$, an $N\times r$ dense matrix A, and a $r\times r \times K$ dense tensor. Besides, we also need to store a sparse tensor $\textbf{M}$ that has the same dimension and sparsity as $\textbf{X}$.

\subsection{Link Prediction}
After the convergence criterion is achieved, we obtain an estimated matrix $\hat{A}$ and an estimated tensor $\hat{\textbf{W}}$. Then the link prediction can be done in the following way.  

For any triplet (i,j,k) that is unobserved in the data $\textbf{X}$, we have  $z_{ijk} = a_i^T W_k a_j +\epsilon$ by the definition of latent variable Z with $\epsilon \sim N(0,1)$, which implies $$z_{ijk} \sim N(a_i^T W_k a_j,1).$$ Thus, $$\hat{p} = Pr(x_{ijk}=1) = Pr(z_{ijk}>0) = \Phi(a_i^T W_k a_j),$$
i.e. the probability of the triplet (i,j,k) being valid is $\Phi(\hat{a}_i^T \hat{W}_k \hat{a}_j)$.

\section{Simulation Study}

\subsection{Simulation Setup}

In this section, we demonstrate the performance of the proposed estimators using the simulated data. We compared the three factorization-based models, including the RESCAL model, the RESCAL\_Logit model and our PTF model.  
Simulation studies with different settings and two generating models are considered. Both the in-sample performance and the out-of-sample performance are evaluated. For the in-sample estimation, we use the canonical correlations between $A$ and $\hat{A}$ to evaluate the parameter estimation. The reason to use the canonical correlations is that the bilinear product $AW_kA^T$ can be written as $(AH)H^{-1}W_kH^{-T}(AH)^T$, which implies the underlying score for a triplet is unchanged when $(A, W_k)$ is replaced by $(AH, H^{-1}W_kH^{-T})$ for any invertible $r\times r$ matrix $H$. For the out-of-sample estimation,
the prediction performances on the unknown relations are measured using AUC with the binary prediction results.
 
The two probabilistic model, the RESCAL\_Logit model and our proposed PTF are used to generate the data.
We first generate entries in $A$  from $N(0, 1)$. The elements in $W_k$ for  $k = 1,... K$ are independently drawn from a Gaussian distribution with variance 1 and means $\mu_k$, which independently follow a $Uniform(-2, -1)$ distribution.  The error term $ \epsilon$ is generated from i.i.d $N(0, 1)$.

\begin{itemize}
\item Model 1: The proposed Probit Tensor Factorization model, where
\begin{center}
$Z_{ijk} = a_i^T W_k a_j + \epsilon, \;\; x_{ijk} = I(Z_{ijk} > 0),$
\end{center}
 for  i = 1,2,...,N, j = 1,2,...,N, k = 1,... K
\item Model 2: The RESCAL\_Logit model  \citep{nickel2013logistic} with
\begin{center}
$ \sigma(Z_{ijk}) = \frac{1}{1+exp(-a_i^T W_k a_j)} + \epsilon, \;\; x_{ijk} \sim Bernoulli(\sigma(Z_{ijk})),$
\end{center}

 \end{itemize}

Table1 summarizes our two simulation settings.  For each simulated data, we randomly set $50\%$,  $70\%$ or $90\%$ of links as missing, which are the triplets to make prediction for.  

\begin{center} 
\begin{table}[]
\centering
\begin{tabular}{c|cccc}
Simulation Setting & N   & K  & r   \\ \hline
1                  & 200 & 10 & 3        \\
2                  & 500 & 20 & 10      \\

\end{tabular}
\caption{Simulation settings. N is the number of entities; K is the number of relationship types; r is the rank for the factorization.}
\label{table:table0}
\end{table}
\end{center}

We compare our proposed model with two other tensor factorization models, which are the RESCAL model \citep[][]{nickel2011three} and RESCAL\_Logit model \citep[][]{nickel2013logistic}.  The results are averaged over 100 repetitions.  We set the maximum number of iterations at 500. The maximum iteration in the M step in our model is set to 3. For RESCAL, we choose a threshold for the output score such that the proportion of zeros and ones in testing match the proportions in training.

\subsection{Simulation results}

In Table \ref{table:table1}, we summarized the performances of the three tensor factorization models when the data is generated from the PTF model. For both in-sample estimation of the latent component matrix $A$ and the out-of-sample link prediction, the PTF model is the best among the three. While all of these models have good performance in in-sample estimation, the PTF model shows its advantage in out-of-sample link prediction. The results are similar in Table \ref{table:table2}, where the RESCAL\_Logit is used as the generating model.

\begin{table}[] \begin{adjustbox}{width=\textwidth}
\begin{tabular}{llllllll}
\multirow{2}{*}{Simulation Setting} & \multirow{2}{*}{Missing Proportion} & \multicolumn{2}{l}{PTF}           & \multicolumn{2}{l}{RESCAL} & \multicolumn{2}{l}{RESCAL\_Logit} \\
                                    &                                     & In-Sample       & Out-Sample      & In-Sample   & Out-Sample   & In-Sample    & Out-Sample         \\ \hline\hline
\multirow{4}{*}{1}        
                                    & 0.5                                  & \textbf{0.997} & \textbf{0.967} & 0.934 & 0.860 &  0.954 &  0.907        \\
                                    & 0.7                                  & \textbf{0.988} & \textbf{0.965} &0.930    &   0.847   &  0.831  &  0.731           \\
                                    & 0.9                                 & \textbf{0.972} & \textbf{0.938} & 0.897   & 0.824  & 0.455     &0.640    \\
                                    \hline
\multirow{4}{*}{2}                
                                    & 0.5                                  & \textbf{0.968} & \textbf{0.987} &  0.948 & 0.896    &  0.966 &   0.957            \\
                                    & 0.7                                  & \textbf{0.967} & \textbf{0.873} &   0.940 &   0.860   & 0.910 & 0.861     \\
                                    & 0.9                                  & \textbf{0.945} & \textbf{0.852} &    0.890   & 0.784  & 0.462  &0.690
\end{tabular}\end{adjustbox}
\caption{In-sample and out-sample performances of the PTF, RESCAL and RESCAL\_Logit model when data is generated from the PTF model. The missing proportion denotes the proportion of unknown triplets in \textbf{X}. The average proportion of zeros is around 50\%.  The in-sample performance is measured by the median canonical correlation of the predicted $A$  and true $A$, while the out-of-sample performance is measured using AUC. The means averaged over 100 repetitions are reported. The standard errors for AUCs are around 0.02.}
\label{table:table1}
\end{table}

\begin{table}[]
\begin{adjustbox}{width=\textwidth}
\begin{tabular}{llllllll}
\multirow{2}{*}{Simulation Setting} & \multirow{2}{*}{Missing Proportion} & \multicolumn{2}{l}{PTF}           & \multicolumn{2}{l}{RESCAL} & \multicolumn{2}{l}{RESCAL\_Logit} \\
                                    &                                     & In-Sample       & Out-Sample      & In-Sample   & Out-Sample   & In-Sample    & Out-Sample         \\ \hline
                                    \hline
\multirow{4}{*}{1}                 
                                    & 0.5           & 0.975 &0.913  & 0.943 & 0.934    &\textbf{0.986}   &\textbf{0.960}      \\
                                    & 0.7           & \textbf{0.946} & \textbf{0.821}   & 0.925      &0.747    & 0.735  &0.661   \\
                                    & 0.9           & \textbf{0.933}  &\textbf{0.768}   & 0.901     &0.743  &  0.122   & 0.520    \\
                                    \hline
\multirow{4}{*}{2}                
                                    & 0.5       & 0.954 & 0.940 &   0.975 &  0.869     &\textbf{0.966}  &\textbf{0.968}   \\
                                    & 0.7        & \textbf{0.941}&  0.822    &0.925       &0.805    &  0.921     & \textbf{0.946}   \\
                                    & 0.9        &\textbf{0.939}  &  \textbf{0.767}     & 0.893      &0.764    & 0.859    & 0.747
\end{tabular}
\end{adjustbox}
\caption{In-sample and out-of-sample performances of the PTF, RESCAL and RESCAL\_Logit model when data is generated from the RESCAL\_Logit model. The in-sample performance is measured by the median canonical correlation of the predicted $A$  and true $A$, while the out-of-sample performance is measured using AUC. The average proportion of zeros is around 50\%. The means averaged over 100 repetitions are reported. The standard errors for AUCs are around 0.02.}
\label{table:table2}
\end{table}

Correct in-sample estimation on the latent component matrices provides opportunities for future quantitative analysis with the entities. For example, it can be used to cluster the entities. The out-of-sample AUC measures how well the models performed in link prediction, which is one of the most common tasks in relational learning. 
From the above two tables, we find that our PTF model achieve good results even when the data is generated from RESCAL\_Logit model. 
While the RESCAL\_Logit model achieved satisfactory prediction performances when the missing proportions are low, it loses its power as the missing proportions increase. One major advantage of the PTF over the other two is that it is able to tell whether a triplet is unknown or invalid thus leverages the most information during training. 
During the simulation, we noticed that there are two other limitations of the RESCAL\_Logit model.  The first one is that the quasi-newton's optimization procedure failed to converge 46\% of the times during the estimation and in general it requires more iterations to converge compared to the other methods. The other one is that the RESCAL\_Logit model has higher computation costs since it needs to compute the dense matrices $AW_kA^T$ in each iteration. The following table displays the average running time per iteration for the PTF model and the RESCAL\_Logit  model, where the PTF model shows its advantage, especially when the missing proportion is high.

\begin{center} 
\begin{table}
\centering
\begin{tabular}{c|ccc|ccc}
Simulation Setting &       & 1     &       &       & 2     &       \\\hline
Missing Proportion & 0.5   & 0.7   & 0.9   & 0.5   & 0.7   & 0.9   \\ \hline
PTF                & \textbf{0.073} & \textbf{0.043} & \textbf{0.013} & 1.415 & \textbf{0.786} & \textbf{0.269} \\
RESCAL\_Logit      & 0.099 & 0.094 & 0.092 & \textbf{1.230} & 1.206 & 1.131
\end{tabular}
\caption{Comparison of running time for the PTF model and the RESCAL\_Logit  model under different setting. The numbers are the seconds that an iteration takes on average over 100 repetitions. }
\end{table}
\end{center}  

We used an information criteria proposed by \cite{shi2019determining} for rank selection, which helps to select the correct rank most of the time during simulation. We are also interested in the model performance when the factorization rank is mis-specified in the PTF model. 
The Table 5 summarized the average AUCs over 100 petitions when rank is mis-specified. Here we set $N$ to 200, $K$ to 10 and missing proportion to $50\%$. Data are generated from the PTF model.
Even when rank is mis-specified, the PTF model still achieves satisfactory performance. 

\begin{center} 
 \begin{table}
 \centering
\begin{tabular}{l|llllllll}
   & 3     & 4     & 5     & 6     & 7     & 8     & 9     & 10    \\ \hline
3  & 0.969 & 0.967 & 0.967 & 0.967 & 0.966 & 0.966 & 0.964 & 0.963 \\
4  & 0.919 & 0.981 & 0.981 & 0.981 & 0.981 & 0.981 & 0.980 & 0.980 \\
5  & 0.915 & 0.947 & 0.988 & 0.988 & 0.988 & 0.988 & 0.988 & 0.988 \\
6  & 0.900 & 0.934 & 0.970 & 0.992 & 0.993 & 0.992 & 0.992 & 0.992 \\
7  & 0.906 & 0.934 & 0.953 & 0.977 & 0.995 & 0.995 & 0.995 & 0.995 \\
8  & 0.883 & 0.906 & 0.929 & 0.952 & 0.976 & 0.995 & 0.995 & 0.995 \\
9  & 0.855 & 0.880 & 0.901 & 0.928 & 0.952 & 0.977 & 0.996 & 0.996 \\
10 & 0.857 & 0.876 & 0.897 & 0.924 & 0.945 & 0.963 & 0.979 & 0.997
\end{tabular}
\caption{The y axis is the true rank used to generate data from the PTF model and the X axis is the rank used to train the model. The numbers are AUC averaged over 100 repetitions.}
\end{table}
\end{center}

\section{Real data application} 
This section presents the results from experiments with three common small- and moderate-sized benchmark datasets: \emph{UMLS}, \emph{Kinships}, \emph{Nations} to evaluate the predictive performance of our model.

\subsection{Datasets}
In order to compare our model with current state-of-art relational learning models, the evaluation was carried out on the following three benchmark datasets:
  
\emph{UMLS}  The data is gathered by \cite{mccray2001aggregating} from the Unified Medical Language System semantic work. It consists 135 entities, which are high-level medical concepts such as \lq Disease or Syndrome\rq, \lq Diagnostic Procedure\rq, or \lq Mammal\rq, and 49 relation types, like \lq affect\rq or \lq cause\rq. There are 6529 valid relations among 893,025 relational facts.

 \emph{Kinships} This dataset describe the Australian kinship system. 
 It was gathered by Denham includes 26 types of relations between 104 tribe members in Alyawarra, a tribe from Central Australia. It describes different kinship relations between members. Examples of kinship relations include \lq Father \rq, \lq Mother \rq, \lq Younger \rq, \lq Female speaker \rq, etc.
  In total there are 10,686 valid triplets out of 281,262 triplets. See  \cite{denham1979aranda} for details.

\emph{Nations} Fifty-six binary relation types between 14 countries are included in this dataset, which results in a 14$\times$14$\times$56 tensor. Out of 12,530 relational facts, 2565 are valid. See \cite{rummel1976dimensionality} for more details. 

There are no missing data in these three datasets, i.e. for each triplet, we know whether it is valid or not. 

\begin{table}[H]
\begin{center}
\begin{tabular}{l|ccc}
  & UMLS & Kinships & Nations\\ \hline
Number of entities (N) & 135  & 104      & 14      \\
Number of relationship types (K) & 49   & 26       & 56     \\
Number of valid triplets (T) & 6529  & 10,686 &  2565 
\end{tabular}
\caption{Summary statistics of the three benchmark datasets.}
\end{center}
\end{table}

\subsection{Experimental Results}

We followed the experimental settings used by \cite{kemp2006learning}. For each of the three datasets, a ten-fold cross-validation was performed with each triplet as one statistical unit. The triplets in the test set are treated as missing. The goal is to accurately predict the missing triplets. Since all the relational facts are known in the datasets, AUC is used to evaluate the prediction accuracy.  During the experiment, the maximum number of iterations in the M step is 2 and the maximum iteration number is 500.  

We consider different types of methods as our competitor. MRC \cite{kok2007statistical} is a Markov-logic based model.
TransE \citep{bordes2013translating} and SME  \citep{bordes2014semantic} are two popular energy-based models.
Results from tensor factorization related models are also reported, including LFM \citep{jenatton2012latent}, RESCAL\_ALS   \citep{nickel2011three}, RESCAL\_Logit \citep{nickel2013logistic}, TATEC  \citep{garcia2016combining}, BPBFM2 \citep{hu2016topic}.
In addition, we include a statistical network analysis model Latentnet \citep{krivitsky2008fitting}, which was applied to each slice of the tensor separately.

Table  \ref{table:table7} summarizes the results of the three experiments. We can see that our proposed PTF model outperforms the competing algorithms on all of these three datasets.  In \emph{UMLS}, AUC from most of competing methods are above 0.980, and our method achieves the highest AUC of 0.998. Regarding the \emph{Kinships} dataset, the performance of the three-way models and various order models perform much better than the two-way models. AUC from  PTF, RESCAL\_Logit, RESCAL are all above 0.95 and the PTF performs the best. In \emph{Nations} dataset, only LFM and PTF achieve AUC higher than 0.9 and the PTF model still performs the best. In summary, in terms of prediction accuracy in these three datasets, the proposed PTF performs uniformly better than the competing methods.

\begin{table}[]
\fontsize{9}{11}\selectfont
\caption{Experimental Results} \label{}
\begin{center}
\scalebox{0.9}{
\begin{tabular}{lllllllll}
Data set& UMLS&Kinships & Nations \\
&N=135, K=49&N=104,K=26&N=14,K=56\\
\hline \\
MRC  \citep{kok2007statistical}     & 0.98                 & 0.84                 & 0.75                 \\
TransE \citep{bordes2013translating}      & 0.734 $\pm$  0.033          & 0.135 $\pm$  0.005          & N/A                   \\
SME  \citep{bordes2014semantic}        & 0.983 $\pm$ 0.003          & 0.907 $\pm$  0.008          & 0.883 $\pm$ 0.02           \\
LFM \citep{jenatton2012latent}   & 0.990 $\pm$0.003  & 0.946 $\pm$  0.005  & 0.909 $\pm$  0.009\\ 
RESCAL\_ALS   \citep{nickel2011three}& 0.98                 & 0.95                 & 0.84                \\
RESCAL\_Logit \citep{nickel2013logistic} & N/A                   & 0.981                & 0.851                \\
TATEC  \citep{garcia2016combining} & 0.985 $\pm$  0.004          & 0.941 $\pm$  0.009           & N/A                   \\
BPBFM2 \citep{hu2016topic} &0.994 &0.976&0.896\\
Latentnet \citep{krivitsky2008fitting}  & 0.511$\pm$  0.009  &0.920 $\pm$  0.003 & 0.840 $\pm$  0.011\\
\textbf{PTF}  & \textbf{0.998 $\pm$  0.002} & \textbf{0.988 $\pm$  0.002} & \textbf{0.928 $\pm$  0.008}
\end{tabular}
}
\end{center}
\label{table:table7}
\end{table}

\section{Discussion} 

In this paper, we proposed a probit tensor factorization model, which is an extension of tensor factorization model for binary data. 
The two parameters in our model, $A$ and $\textbf{W}$, correspond to the low-rank representation of entities and characterization of different relations. 
We illustrate our model fitting procedure under the framework of EM algorithm. Scalability issue is addressed by showing that the complexity of our algorithm is linear in the number of entities, the number of known facts and the number of relationship types. We conducted experiments on three common medium sized datasets, where our method obtains uniformly better results than the other state-of-art models. 
One major advantage of our model is that it adopts the open world assumption (OWA) and allows a natural use of invalid facts in model. And it outputs an interpretable probabilistic result that can be further used in quantitative analysis.  

While our model is designed to deal with asymmetric relations, it can also be used to perform link predictions for symmetric relations. 
We need to impose a constraint that the latent representation for relations $W_k$'s are symmetric metric. With the constraint, the update rule for $W_k$ can be formulated as a quadratic programming problem.


Another interesting topic is how to decide the factorization rank, where both computation cost and predictive performance should be taken into consideration. In the literature,
\cite{mazumder2010spectral} applied regularization to select the rank for singular value decomposition,
\cite{nickel2011three} suggested to use cross-validation for rank selection in RESCAL model
and
\cite{sun2017provable} proposed to use Bayesian information criteria (BIC) \citep{schwarz1978estimating} for sparse CP decomposition.
Recently, 
\cite{shi2019determining} proposed a general class of information criteria to determine the rank for RESCAL model and showed their method achieved good finite sample properties. The proposed method is also applicable to our model and we found it works well in our simulation. 

The PTF model can be extended in several directions. One way is to incorporate prior information. For example, by utilizing prior domain knowledge of type constraints that is suggested by \cite{chang2014typed}, we can further improve the efficiency. And more prior information, e.g.  logical structure of the relations, attributes of the entities and so on are expected to be included in tensor factorization model to further enhance the model performance. 

\appendix
\section{Appendix}

\subsection{Runtime for the EM algorithm}
We summarized runtime for both E step and M step with different entity size $N$, relation size $K$, number of observed triplets $T$ and decomposition rank $r$.  Note that the computation complexity  is $O(Tr^2)$  for each E-step and $O(NKr^2 + Tr^2)$ for each M-step. 
All runtimes are averaged over 100 repetitions. The runtime performance is aligned with our expectation.

\begin{table}[H]
\caption{Runtime Table} \label{}
\begin{adjustbox}{width=\textwidth}
\begin{tabular}{l|llll|cc}
\hline
Varying parameters & $N$       & $K$   &$ r $ &$ T$        & Runtime for E step (seconds) & Runtime for M step (seconds) \\ \hline
$N$                & 1,000   & 50  & 4  & 10,000   & 0.00754            & 0.0213             \\
                   & 10,000  & 50  & 4  & 10,000   & 0.00758            & 0.207              \\
                   & 100,000 & 50  & 4  & 10,000   & 0.0114             & 2.11               \\ \hline
$K$                & 1,000   & 50  & 4  & 10,000   & 0.00754            & 0.0213             \\
                   & 1,000   & 100 & 4  & 10,000   & 0.00842            & 0.0498             \\
                   & 1,000   & 200 & 4  & 10,000   & 0.00842            & 0.0906             \\ \hline
$r$                & 1,000   & 50  & 4  & 10,000   & 0.00754            & 0.0213             \\
                   & 1,000   & 50  & 8  & 10,000   & 0.0121             & 0.0685             \\
                   & 1,000   & 50  & 16 & 10,000   & 0.0396             & 0.558              \\ \hline
$T$                & 1,000   & 50  & 4  & 10,000   & 0.00754            & 0.0213             \\
                   & 1,000   & 50  & 4  & 100,000  & 0.0646             & 0.0758             \\
                   & 1,000   & 50  & 4  & 1000,000 & 0.626              & 0.530              \\ \hline
\end{tabular}
\end{adjustbox}
\end{table}

\bibliographystyle{Chicago}

\bibliography{bibliograph}
\end{document}